\documentclass[journal]{IEEEtran}
\usepackage{hyperref}
\hypersetup{
	colorlinks=true,
	linkcolor=black,     
	urlcolor=black,
	citecolor = black
}
\usepackage{bbm}
\usepackage{caption}
\usepackage{amsthm}
\usepackage{utfsym}
\usepackage{amssymb,graphicx,lineno}
\usepackage{rawfonts,amsfonts,amssymb,psfrag,color}
\usepackage{amsfonts}
\usepackage{amsmath,epsfig}
\usepackage{rawfonts,graphicx,amsfonts,amssymb}
\usepackage{bm}
\usepackage{algorithm, algorithmic}
\usepackage{subfig}
\usepackage{graphicx}
\usepackage{float}
\usepackage{verbatim}
\usepackage{color,xcolor}

\newtheorem{theorem}{Theorem}

\modulolinenumbers[5]
\usepackage{multirow}
\usepackage{booktabs}
\renewcommand{\arraystretch}{1.5}

\usepackage{cite}
\usepackage{color}

\usepackage{comment}

\begin{document}
	\title{Robust matrix completion via Novel M-estimator Functions }
	%
	\author{Zhi-Yong Wang and Hing Cheung So,~\IEEEmembership{Fellow,~IEEE} \thanks{E-mail: z.y.wang@my.cityu.edu.hk, hcso@ee.cityu.edu.hk. This work is supported by a grant from the Research Grants Council of the Hong Kong Special Administrative Region, China [Project No. CityU 11207922]. }}

	\maketitle
	\begin{abstract}
		M-estmators including the Welsch and Cauchy have been widely adopted for robustness against outliers, but they also down-weigh the uncontaminated data. 
		To address this issue, we devise a framework to generate a class of nonconvex functions which only down-weigh outlier-corrupted observations.  
		Our framework is then applied to the Welsch, Cauchy and $\ell_p$-norm functions to produce the corresponding robust loss functions. 
		Targeting on the application of robust matrix completion, efficient algorithms based on these functions are developed and their convergence is analyzed. 
		Finally, extensive numerical results demonstrate that the proposed methods are superior to the competitors in terms of recovery accuracy and runtime.
	\end{abstract}
	\section{Introduction}
	\label{sec:intro}
	
	Matrix completion (MC)~\cite{Davenport2016, YChenY2018} refers to recovering the missing entries of a partially-observed matrix. It has numerous applications in signal processing and machine learning, such as hyperspectral imaging~\cite{WangZYRTCSVT2023} and image inpainting~\cite{WangZYATCSVT2023}. 
	MC can be formulated as a constrained rank minimization problem~\cite{Cand2009}, but it is NP-hard since the rank is discrete. To address this, nuclear norm minimization is exploited~\cite{Fazel2002} to recast MC as a semi-definite program~\cite{Cand2010}. Although it can be solved by the interior-point method~\cite{LiuZ2009}, its computational complexity is high. On the other hand, computationally efficient algorithms such as singular value thresholding~\cite{Cai2010}, and accelerated proximal gradient with linesearch~\cite{Toh2010}, have been proposed. Nevertheless, they still involve full singular value decomposition (SVD) per iteration, implying an expensive cost especially for large-size data. 
	
	To avoid performing SVD, factorization based MC approach~\cite{ZhuZ2018, Zhu2021} has been developed, whose idea is to utilize the product of two much smaller matrices to approximately match the observed matrix and thus the low-rank property is automatically fulfilled. MC algorithms such as low-rank matrix fitting~\cite{WenZS2012} and alternating minimization~\cite{JainPL2013}, are then proposed. In addition, Zhu $et$ $al$.~\cite{ZhuZ2018, Zhu2021} have shown that this kind of MC problem has no spurious local minima and obeys the strict saddle property which requires the cost function to have a directional negative curvature at all critical points but local minima. That is, although the factorization based MC problem is nonconvex, global optimality can be achieved under some conditions. Nevertheless, the above mentioned methods are vulnerable to outliers. 
	
	To resist outliers, $\ell_1$-norm is suggested, resulting in numerous robust MC algorithms, including robust matrix	factorization by majorization minimization (RMF-MM)~\cite{LinZR2018} and practical low-rank matrix approximation under robust $\ell_1$-norm (RegL$_1$)~\cite{ZhengYP2012}. Nevertheless, the $\ell_1$-norm is still vulnerable to outliers with large magnitudes because it is not upper bounded~\cite{ZengWJO2018}. To enhance outlier-robustness, the $\ell_p$-norm with $0<p<1$~\cite{ZengWJO2018,ShangFB2018} and the nonconvex penalty functions such as the Welsch and Tukey, are adopted~\cite{HeYR2019,MumaMR2019}. Although these functions penalize outliers via assigning small weights to outlier-corrupted data, they also down-weigh the normal data. Here, normal data refer to observations without noise or with only Gaussian noise. Compared with these nonconvex functions, the Huber function only penalizes outlier-contaminated entries but is still sensitive to large outliers because it employs the $\ell_1$-norm for robustness~\cite{WangZYRTSP2023}.
	
	In this paper, we devise a framework to produce a class of M-estimator functions, which only down-weigh outlier-contaminated data and can resist even large outliers. The framework is then applied to commonly-used Welsch, Cauchy and $\ell_p$-norm M-estimators, resulting in the corresponding robust functions. Since these functions are nonconvex, the Legendre-Fenchel (LF) transform is exploited to convert the nonconvex problem into a sum of convex problems with closed-form solutions. Furthermore, we apply the developed functions to factorization based MC and propose robust MC algorithms with convergence guarantees.

	

	\section{Problem formulation}
	\label{sec:pro-for}
	
	Let ${\Omega} \subset \{1,\cdots,m\}\times\{1,\cdots,n\}$ represent the index set of the known entries of an incomplete matrix $\pmb X_\Omega$, and $(\cdot)_{\Omega}$ is a projection operator, defined as:
	$$ \left[\pmb X_{\Omega}\right]_{ij}  = \left\{
	\begin{aligned}
		& X_{ij},\quad   {\rm if}~(i,j)\in \Omega  \\
		&0,\quad \quad           \rm {otherwise}.  \\
	\end{aligned}
	\right.
	$$
	Given $\pmb X_\Omega$, MC is to seek a low-rank matrix $\pmb M$ to match $\pmb X_\Omega$ and estimate its missing entries, which can be modeled as a rank minimization problem~\cite{Cand2009}:
	\begin{equation}\label{NP-hard}
		\begin{split}
			&\mathop {\min}\limits_{\pmb M}~ \text{rank}(\pmb M), ~\text{s.t.} ~ \pmb M_{\Omega} = \pmb X_{\Omega}
		\end{split}    	   	
	\end{equation}
	However, (\ref{NP-hard}) is an NP-hard problem. Instead, the convex nuclear norm is suggested~\cite{Fazel2002}:
	\begin{equation}\label{nuclear-norm}
		\begin{split}
			&\mathop {\min}\limits_{\pmb M}~ \|{\pmb M}\|_*, ~ \text{s.t.} ~ \pmb M_{\Omega} = \pmb X_{\Omega}
		\end{split}    	   	
	\end{equation}
	where the nuclear norm $\|\pmb M\|_*$ is the sum of singular values of $\pmb M$. Although it is convex, full SVD calculation is required per iteration. To address this problem, factorization based MC has been exploited~\cite{WenZS2012}:
	\begin{equation}\label{matrix-factorization}
		\begin{split}
			&\mathop {\min}\limits_{\pmb {U,V}}~ \left \|{\pmb X_\Omega} - \left(\pmb U\pmb V\right)_\Omega\right \|_F^2
		\end{split}    	   	
	\end{equation}
	where $\pmb U \in \mathbb{R}^{m\times r}$ and $\pmb V \in \mathbb{R}^{r\times n}$ are the two small size matrices with rank $r\ll {\rm min}(m,n)$. Although the restored matrix $\pmb M = \pmb U \pmb V$ is low-rank, 
	the recovery performance is sensitive to outliers because the $\ell_2$-norm is utilized. 
	
	To resist outliers, loss functions such as the Huber and Welsch functions are suggested~\cite{HeYR2019,MumaMR2019}, resulting in:
		\begin{equation}\label{robust_matrix-factorization}
		\begin{split}
			&\mathop {\min}\limits_{\pmb {U,V}}~ l({\pmb X_\Omega} - \left(\pmb U\pmb V\right)_\Omega)
		\end{split}    	   	
	\end{equation}
	where $l(\cdot)$ refers to a robust loss function. Table~\ref{weight-funs} tabulates the commonly-used loss functions and the corresponding weight functions. It is known that a good robust loss function should only down-weigh the outlier-corrupted elements. We see that the quadratic function, namely, the $\ell_2$-norm, has the same weights for all entries, including the outlier-contaminated data, thus it is suitable to handle normal data but is not robust against outliers. Although the Huber function only assigns small weights for noisy observations, it is still vulnerable to large outliers since it employs the $\ell_1$-norm to combat gross errors. To enhance outlier-robustness, the nonconvex functions such as the Welsch and Cauchy, are suggested~\cite{MandanasFD2017}. However, as shown in Table~\ref{weight-funs}, these functions also reduce the weights of normal data. Recently, we design a novel robust function called hybrid ordinary-Welsch (HOW)~\cite{WangZYRTSP2023} where `ordinary' refers to the quadratic function, whose expression is shown in Table~\ref{weight-funs}. We see that only the Huber and HOW functions down-weigh outlier-corrupted entries because they assign the same weights for $|x|\leq c$, where $c$ is a parameter to differentiate whether an entry is contaminated by outliers or not. That is, when $|x|>c$, the corresponding element is considered corrupted by an outlier, and assumed normal entry otherwise.
	
	
	\begin{table*}[htb]
		\renewcommand\arraystretch{1}   
		\caption{\small {Commonly-used loss functions and their weight functions}} 
		\vspace{-0.5cm}
		\begin{center}
			\setlength{\tabcolsep}{0.9mm}{
				{\begin{tabular}{ |c|c|c|c|c|c|}
						\hline
						& Quadratic &Huber & Cauchy & Welsch & HOW \\
						\hline
						$l(x)$& $\frac{x^2}{2}$ &$\begin{cases}
							x^2/2, &|x|\leq c\\
							c|x|-\frac{c^2}{2}, &|x|\textgreater c
						\end{cases}$ & $\frac{\gamma^2}{2}\ln \left(1+\left(\frac{x}{\gamma}\right)^2\right)$& $\frac{\sigma^2}{2}\left(1-e^{\frac{-x^2}{\sigma^2}}\right)$ & $\begin{cases}
							x^2/2, &|x|\leq c\\
							\frac{\sigma^2}{2}\left(1-e^{\frac{c^2-x^2}{\sigma^2}}\right)+\frac{c^2}{2}, &|x|\textgreater c
						\end{cases}$ \\
						\hline
						$w(x)=\frac{l'(x)}{x}$& $1$ &$\begin{cases}
							1, &|x|\leq c\\
							c\cdot{\rm sign}(x)/x, &|x|\textgreater c
						\end{cases}$ &$\frac{1}{1+x^2/\gamma^2}$&$e^{\frac{-x^2}{\sigma^2}}$ & $\begin{cases}
							1, &|x|\leq c\\
							e^{\frac{c^2-x^2}{\sigma^2}}, &|x|\textgreater c
						\end{cases}$ \\
						\hline
			\end{tabular}}}
			\label{weight-funs}
		\end{center}
		\vspace{-0.5em}
	\end{table*}

	\section{Novel M-estimator function and its application to robust matrix completion}
	\label{sec:BVP}
	
	In this section, motivated by the thoughts of the construction for Huber and HOW, we devise a framework to generate a class of M-estimator functions, which only down-weights the outlier-corrupted data. 

	\subsection{Framework to Generate M-estimator Functions}
	\label{R-OR1MP}
	
	We generalize the expressions of Huber and HOW functions for $|x|>c$ and develop a new generic function:
	\begin{equation}\label{Def_framework}
		l_{g,c}(x) = 
		\begin{cases}
			x^2/2, &|x|\leq c\\
			a\cdot g(|x|)+b, &|x|\textgreater c
		\end{cases}
	\end{equation}
	where $g(x)$ is a continuous function and $g'(x)\geq 0$ for $x>0$, while $a$ and $b$ are constants to ensure that $l_{g,c}(x)$ is continuous and smooth at $x=c$. Thus, $a=c/g'(c)>0$ ($g'(c)\neq 0$), and $b=c^2/2-ag(c)$.
	We easily see that $l_{g,c}$ only down-weighs the outlier-corrupted data because when $|x|\leq c$, $l_{g,c}$ is the quadratic function and does not reduce the weight of normal data, while when $|x|\textgreater c$, $l_{g,c}$ employs the robust function $g$ to handle outliers. Compared with the Huber function where $g(x)=|x|_1$, we will focus on nonconvex $g(x)$ because the latter can resist large outliers. Hence, the function $l_{g,c}$ is assumed nonconvex in our study.
	
	Analogous to~\cite{WangZYRTSP2023}, the LF transform~\cite{RockafellarRT2004} is applied to $l_{g,c}$, resulting in:
	\begin{equation}\label{l-new-function}
		\begin{split}
			l_{g,c}(x) = \mathop {\inf}\limits_{y} ~\frac{(y-x)^2}{2} + \varphi_{g,c}(y)
		\end{split}	
	\end{equation}	
	where $\varphi_{g,c}$ is the dual function of $l_{g,c}$ and is also called the implicit regularizer (IR).
	The value of $y$ that solves (\ref{l-new-function}) is:
	\begin{equation}\label{y-pro-solution}
		P_{\varphi_{g,c}}(x) := \nabla f(x) = {\rm max}\left\{0, |x|-a\cdot g'(|x|) \right\}\cdot {\rm sign}(x)
	\end{equation}
	The process of obtaining (\ref{y-pro-solution}) from (\ref{Def_framework}) can be found in~\cite{WangZYRTSP2023}.
	
	Next, we will specify $g$ as several commonly-used functions, including the $\ell_p$-norm and Cauchy function. Note that we have already specified $g$ as the Welsch M-estimator and have proposed the HOW function in~\cite{WangZYRTSP2023}.
	
	When $g(x)=|x|^p$, we have the smooth hybrid ordinary-$\ell_p$ (HOP) function:
	\begin{equation}\label{lp-ab}
		l_{p,c}(x) = 
		\begin{cases}
			x^2/2, &|x|\leq c\\
			\frac{1}{p}c^{2-p} |x|^p+\frac{c^2}{2}-\frac{1}{p}c^2, &|x|\textgreater c
		\end{cases}
	\end{equation}
	By the LF transform, we obtain
	\begin{equation}\label{l-p-function}
		\begin{split}
			l_{p,c}(x) = \mathop {\min}\limits_{y} ~\frac{(x-y)^2}{2} + \varphi_{p,c}(y)
		\end{split}	
	\end{equation}
	where $\varphi_{p,c}(y) $ is the IR related to $l_{p,c}(x)$,	and the solution to $y$ is:
	\begin{equation}\label{lp-pro-solution}
		P_{{\varphi}_{p,c}}(x) = {\rm max}\left\{0, |x|-{c}^{2-p}|x|^{p-1} \right\}\cdot {\rm sign}(x)
	\end{equation}
	Note that when $p=1$, (\ref{lp-ab}) becomes the Huber function.
	
	We then replace $g(x)$ by the Cauchy M-estimator shown in Table~\ref{weight-funs}, and then develop the hybrid ordinary-Cauchy (HOC) function:
	\begin{equation*}
		l_{\gamma,c}(x) = 
		\begin{cases}
			x^2/2, &|x|\leq c\\
			\frac{\gamma^2+c^2}{2}\ln \left(1+\left(\frac{x}{\gamma}\right)^2\right)+b, &|x|\textgreater c
		\end{cases}
	\end{equation*}
	where $b=\frac{c^2}{2}-\frac{\gamma^2+c^2}{2}\ln \left(1+\left({c}/{\gamma}\right)^2\right)$ and $\gamma$ is the scale parameter. Employing the LF transform results in:
	\begin{equation*}
		\begin{split}
			l_{\gamma,c}(x) = \mathop {\min}\limits_{y} ~\frac{(x-y)^2}{2} + \varphi_{\gamma,c}(y)
		\end{split}	
	\end{equation*}
	with the solution to $y$ being:
	\begin{equation}\label{lc-pro-solution}
		p_{\varphi_{\gamma,c}}(x) := {\rm max}\left\{0, |x|-\frac{\left(\gamma^2+c^2\right)|x|}{\gamma^2+x^2}\right\}\cdot {\rm sign}(x)
	\end{equation}

	\subsection{Algorithms for Robust Matrix Completion}
	
	We replace the Frobenius norm in (\ref{matrix-factorization}) with our M-estimator functions, leading to:
	
	\begin{equation}\label{RMC-l-loss}
		\mathop {\min}\limits_{\pmb U,\pmb V}~l_{g,c}\left(\pmb X_\Omega -  \left(\pmb U \pmb V\right)_\Omega\right)
	\end{equation}
	where $l_{g,c}\left(\pmb X_\Omega -  \left(\pmb U \pmb V\right)_\Omega\right)$ is separable, i.e., $l_{g,c}\left(\pmb X_\Omega -  \left(\pmb U \pmb V\right)_\Omega\right) \\= \sum_{i,j\in \Omega} l_{g,c}\left(\pmb X_{i,j} -  \left(\pmb U \pmb V\right)_{i,j}\right)$.
	According to (\ref{l-new-function}), we have:
	\begin{equation}\label{RMC-l-loss-equal}
		\mathop {\min}\limits_{\pmb U,\pmb V, \pmb S} ~\mathcal{L}_{g,c}\left(\pmb U,\pmb V,\pmb S\right) :=\frac{1}{2}\left\|\pmb X_\Omega -  \left(\pmb U \pmb V\right)_\Omega - \pmb S_\Omega \right\|_F^2 + \varphi_{g,c}(\pmb S_\Omega)
	\end{equation}
	where $\varphi(\pmb S) = \sum_{i,j} \varphi(\pmb S_{i,j})$, and $\pmb S_{\Omega^c} = \pmb 0$.
	In the $(k+1)$th iteration, given $\pmb U^k$ and $\pmb V^k$, (\ref{RMC-l-loss-equal}) is equal to:
	\begin{equation}\label{S-problem}
		\mathop {\min}\limits_{\pmb S} ~\frac{1}{2}\left\|\pmb D^k_\Omega - \pmb S_\Omega \right\|_F^2 + \varphi_{g,c}(\pmb S_\Omega)
	\end{equation}
	where $\pmb D^k = \pmb X -  \pmb U^k \pmb V^k$. The solution to (\ref{S-problem}) via (\ref{y-pro-solution}) is:
	\begin{equation}\label{S-solution}
		\pmb S^{k+1}_\Omega = { P}_{\varphi_{g,c}}(\pmb D^k_\Omega )
	\end{equation}
	Given $\pmb S^{k+1}$, (\ref{RMC-l-loss-equal}) amounts to:
	\begin{equation}\label{UV-problem}
		\mathop {\min}\limits_{\pmb U,\pmb V} ~h(\pmb U,\pmb V):=\frac{1}{2}\left\|\pmb H^{k+1}_\Omega -  \left(\pmb U \pmb V\right)_\Omega\right\|_F^2
	\end{equation}
	where $\pmb H^{k+1}_\Omega = \pmb X_\Omega - \pmb S^{k+1}_\Omega$, and it can be efficiently solved by the scaled alternating steepest descent (SASD)~\cite{TannerJ2015}.
	Then, the scaled gradient descent directions for our case are:
	\begin{subequations}
		\begin{align}
			&\widetilde{\nabla} h_{\pmb V}(\pmb U) = \left(\pmb H_\Omega - (\pmb U \pmb V)_\Omega\right)\pmb V^T(\pmb V \pmb V^T)^{-1}\label{S-gradient-U}\\
			&\widetilde{\nabla} h_{\pmb U}(\pmb V) = (\pmb U^T \pmb U)^{-1}\pmb U^T\left(\pmb H_\Omega - (\pmb U \pmb V)_\Omega\right)\label{S-gradient-V}
		\end{align}
	\end{subequations}
	with the corresponding step sizes being:
	\begin{subequations}
		\begin{align}
			&\widetilde{\mu}_{\pmb U} = \left< \nabla h_{\pmb V}(\pmb U), \widetilde{\nabla} h_{\pmb V}(\pmb U)\right>\bigg/\left\| \left(\widetilde{\nabla} h_{\pmb V}(\pmb U)\pmb V\right)_\Omega\right\|_F^2\label{S-step-mu-U}\\
			&\widetilde{\mu}_{\pmb V} =  \left< \nabla h_{\pmb U}(\pmb V), \widetilde{\nabla} h_{\pmb U}(\pmb V)\right>\bigg/\left\| \left(\pmb U \widetilde{\nabla} h_{\pmb U}(\pmb V)\right)_\Omega\right\|_F^2\label{S-step-mu-V}
		\end{align}
	\end{subequations}
	Thus, the SASD updates for $\pmb U$ and $\pmb V$ are:
	\begin{subequations}
		\begin{align}
			&\pmb U^{k+1} = \pmb U^k - \widetilde{\mu}_{\pmb U}^{k}\widetilde{\nabla} h_{\pmb V^k}(\pmb U^k)\label{S-update-U}\\
			&\pmb V^{k+1} = \pmb V^k - \widetilde{\mu}_{\pmb V}^{k}\widetilde{\nabla} h_{\pmb U^{k+1}}(\pmb V^k) \label{S-update-V}
		\end{align}
	\end{subequations}
	
	Recall that the value of $c$ in (\ref{Def_framework}) is the boundary to determine whether an entry is corrupted or not. Similar to~\cite{HeYR2019,WangZYRTSP2023}, its value is set as:
	\begin{equation}\label{choosing-sig}
		\begin{split}
			c^k &= \min\left\{ \xi d^k,c^{k-1}\right\}
		\end{split}    	   	
	\end{equation}
	where $\xi \textgreater 0$ is a user-defined constant, $d^k$ is the robust normalized interquartile range of the vectorized $\pmb D_\Omega$, defined as:
	\begin{equation}\label{choosing-sigma}
		\begin{split}
			d^k = {\rm IQR}\left({\rm vec}(\pmb D_\Omega^k)\right)/1.349
		\end{split}    	   	
	\end{equation}
	with ${\rm IQR}(\cdot)$ being the sample interquartile range operator~\cite{MaronnaRA2006}.
	
	When HOW, HOP and HOC are adopted in (\ref{RMC-l-loss}), the resultant algorithms are referred to as robust MC via HOW (RMC-HOW), HOP (RMC-HOP) and HOC (RMC-HOC), respectively. 
	
	In addition, SASD dominates the complexity of the proposed approach with complexity of $\mathcal{O}\left(8|\Omega|r+4(m+n)r^2\right)$ per iteration. Defining $E_k = l_{g,c}\left(\pmb X_\Omega -  \left(\pmb U^k \pmb V^k\right)_\Omega\right)$, we terminate our algorithms until the relative error $rel_E^k = \left(E_{k}-E_{k-1}\right)/E_{k-1}<\zeta$. On the other hand, the convergence analysis results are shown in the following theorems, whose proofs are analogous to those in our previous work~\cite{WangZYRTSP2023}, and we omit them due to page limit.
	\begin{theorem}\label{C-convergence} 
		The generated sequence $\left\{\mathcal{L}_{g^k,c^k}\left(\pmb U^{k},\pmb V^{k},\pmb S^{k}\right)\right\}$ converges.
	\end{theorem}
	\begin{theorem}\label{C-critical-convergence} 
		Let $\left\{{\left(\pmb U^k,\pmb V^k,\pmb S^k\right)}\right\}$ be the generated sequence, and suppose that $\left(\pmb U^k,\pmb V^k\right)$ are of full rank. Then, $\left\{{\left(\pmb U^k,\pmb V^k,\pmb S^k\right)}\right\}$ is bounded. Besides, let $\left\{{\left(\pmb U^{k_j},~\pmb V^{k_j},~\pmb S^{k_j}\right)}\right\}$ be a generated subsequence such that $\lim_{k_j\rightarrow \infty} \left(\pmb U^{k_j},\pmb V^{k_j},\pmb S^{k_j}\right)\\ = \left(\pmb U^\star,\pmb V^\star,\pmb S^\star\right)$. Then, ${\left(\pmb U^\star,\pmb V^\star,\pmb S^\star\right)}$ is a critical point.
	\end{theorem}

	\section{Experimental Results}
	\label{sec:sim}
	
	We compare our algorithms with the competing methods, including HQ-ASD~\cite{HeYR2019}, $(\pmb S+\pmb L)_{{1}/{2}}$~\cite{ShangFB2018}, $(\pmb S+\pmb L)_{{2}/{3}}$~\cite{ShangFB2018}, RMF-MM~\cite{LinZR2018} and $\rm RegL_1$~\cite{ZhengYP2012}. All numerical simulations are conducted using a computer with 3.0 GHz CPU and 16 GB memory. We first generate a low-rank matrix $\pmb X = \pmb U \pmb V$, where the entries of $\pmb U \in \mathbb{R}^{m\times r}$ and $\pmb V \in \mathbb{R}^{r\times n}$ satisfy the standard Gaussian distribution. Impulsive noise is modeled by the Gaussian mixture model (GMM) while the signal-to-noise ratio (SNR) is defined as:
	\begin{equation}
		{\rm SNR} = \frac{\|\pmb X_\Omega\|_F^2}{|\Omega|\left((1-\tau)\sigma_1^2+\tau\sigma_2^2\right)}
	\end{equation}
	where $\sigma_1^2$ and $\sigma_2^2$ are variances with $\sigma_1^2 \ll \sigma_2^2$, and $\tau$ controls the proportion of outliers. To model outliers, we set $\sigma_2^2=100\sigma_1^2$ and $\tau=0.1$. 
	Besides, the root mean square error (RMSE) defined as ${\rm RMSE} = {\|\pmb X - \pmb M\|_F}/{\sqrt{mn}}$ is utilized to measure the performance of all algorithms. Similar to the parameter selection in~\cite{WangZYRTSP2023}, we set $\xi=2$ and $\zeta=10^{-4}$ in our methods.
	
	\begin{figure}[htb]
		\centering
		\includegraphics[width=6cm]{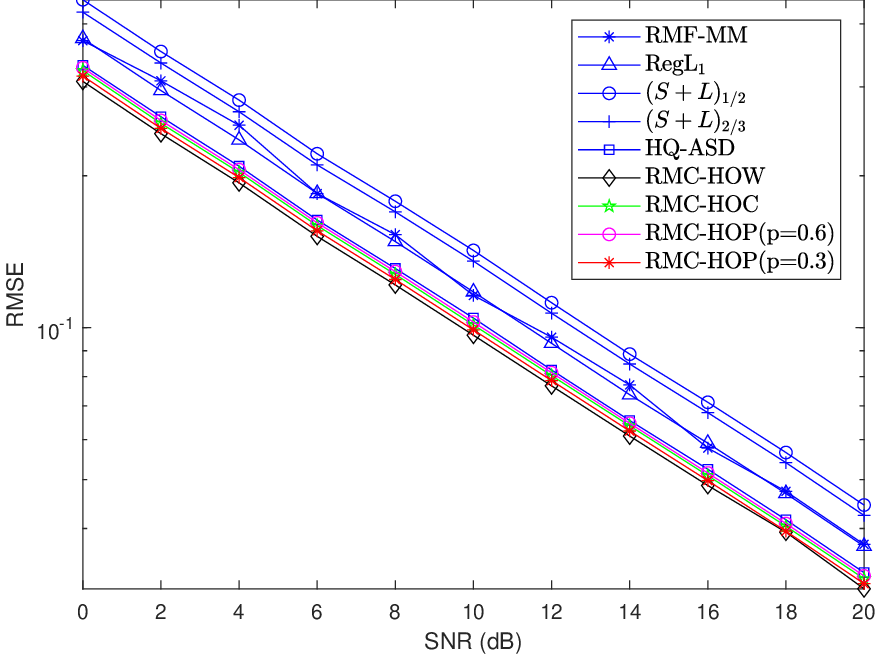}
		\vspace{-0.5em}
		\caption{RMSE versus SNR.}\label{RMSE_SNR}
	\end{figure}
	
	\begin{figure}[htb]
		\centering
		\includegraphics[width=6cm]{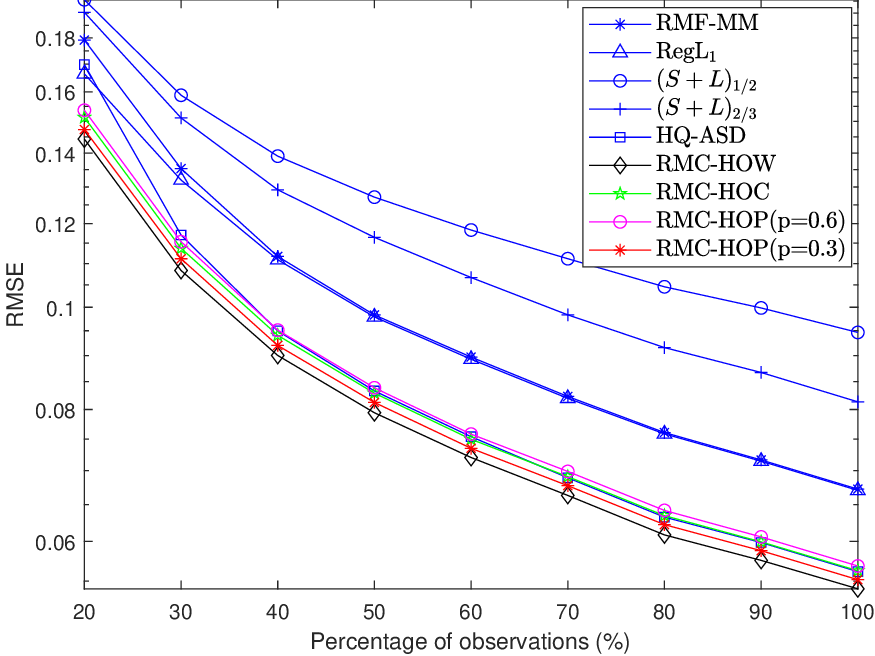}
		\vspace{-0.5em}
		\caption{RMSE versus percentage of observations.}\label{RMSE_per}
	\end{figure}

	\begin{table}[htb]
		\renewcommand\arraystretch{1}   
		\caption{\small {Average runtime (in seconds) for different matrix sizes. }}  
		\begin{center}
			\setlength{\tabcolsep}{1.2mm}{
				\begin{tabular}{ccccc}
					\hline
					Method& Case 1& Case 2& Case 3& Case 4\\
					\hline
					RMF-MM	& \multicolumn{1}{c}{$22.98$} &\multicolumn{1}{c}{$175.1$} & \multicolumn{1}{c}{$460.7$}   &\multicolumn{1}{c}{$855.2 $} \\
				RegL$_1$	& \multicolumn{1}{c}{$0.161$} &\multicolumn{1}{c}{$0.963$} & \multicolumn{1}{c}{$2.233$}   &\multicolumn{1}{c}{$5.346 $} \\
			$(S+L)_{1/2}$	& \multicolumn{1}{c}{$ 0.663$} &\multicolumn{1}{c}{$ 2.548$} & \multicolumn{1}{c}{$6.349 $}   &\multicolumn{1}{c}{$13.30  $} \\
			$(S+L)_{2/3}$	& \multicolumn{1}{c}{$0.573 $} &\multicolumn{1}{c}{$2.340 $} & \multicolumn{1}{c}{$5.852 $}   &\multicolumn{1}{c}{$11.59  $} \\
					HQ-ASD	& \multicolumn{1}{c}{$0.080 $} &\multicolumn{1}{c}{$ 0.372$} & \multicolumn{1}{c}{$ 0.846$}   &\multicolumn{1}{c}{$1.765  $} \\
					RMC-HOW	& \multicolumn{1}{c}{$0.027 $} &\multicolumn{1}{c}{$0.118 $} & \multicolumn{1}{c}{$ 0.268$}   &\multicolumn{1}{c}{$0.532  $} \\
					RMC-HOC	& \multicolumn{1}{c}{$ 0.029$} &\multicolumn{1}{c}{$ 0.136$} & \multicolumn{1}{c}{$0.300 $}   &\multicolumn{1}{c}{$ 0.608 $} \\
		RMC-HOP ($p=0.6$)	& \multicolumn{1}{c}{$ 0.048$} &\multicolumn{1}{c}{$ 0.199$} & \multicolumn{1}{c}{$0.425 $}   &\multicolumn{1}{c}{$ 0.812 $} \\
		RMC-HOP ($p=0.3$)	& \multicolumn{1}{c}{$0.050 $} &\multicolumn{1}{c}{$0.205 $} & \multicolumn{1}{c}{$0.454 $}   &\multicolumn{1}{c}{$  0.864$} \\
					\hline
			\end{tabular}}
			\vspace{-2em}
			\label{RMC-runtime}
		\end{center}
	\end{table}
	
	We conduct experiments on data matrices with $m=300$, $n=200$ and $r=5$. Fig.~\ref{RMSE_SNR} plots the RMSE versus SNR with $30\%$ observations. It is seen that the RMSE for all methods decreases with SNR, while the proposed algorithms have smaller recovery error than the competing techniques. In addition, RMC-HOW yields the best recovery because HOW is bounded. Moreover, the impact of percentage of observations is investigated, and the results are shown in Fig.~\ref{RMSE_per}. Again, our methods outperform the competitors, with HOW attaining the best recovery performance. 
	
	Finally, we investigate the runtime of all methods, and perform numerical experiments on four cases with different matrix dimensions. Here, Case 1: $m=300,n=200,r=5$, Case 2: $m=600,n=400,r=10$, Case 3: $m=900,n=600,r=15$, and Case 4: $m=1200,n=800,r=20$. The results under SNR = $10$ dB and $50\%$ observations are tabulated in Table~\ref{RMC-runtime}. We see that the runtime of our algorithms is less than that of the competitors.

	\section{Conclusion}
	\label{sec:con}
	
	In this paper, we provide a framework to generate a new class of robust loss functions via combining the quadratic and other robust loss functions. The proposed functions can be used to combat gross errors and only down-weigh the outlier-contaminated observations. Applying our framework to the Welsch, Cauchy and $\ell_p$-norm functions yields the HOW, HOC and HOP, respectively, which are then adopted for robust MC. Furthermore, although the resultant optimization problem is nonconvex, the LF transform is adopted to transform it into a sum of convex subproblems. 
	Then, efficient robust MC algorithms based on SASD are developed.
	Finally, experimental results show the superiority of the proposed algorithms over the competing methods in terms of recovery error and runtime.
	
		
	\bibliographystyle{IEEEbib}
	\bibliography{strings,refs}

\end{document}